\documentclass[10pt]{article}
\usepackage[preprint]{colm2025_conference}

\usepackage{microtype}
\usepackage{hyperref}
\usepackage{url}
\usepackage{booktabs}
\usepackage{float}
\usepackage{graphicx}
\usepackage{lineno}
\usepackage{amsmath}
\usepackage{algorithm}

\usepackage{algpseudocode}
\usepackage{enumitem}
\usepackage{svg}
\setlist[itemize]{leftmargin=*, nosep}

\definecolor{darkblue}{rgb}{0, 0, 0.5}
\hypersetup{colorlinks=true, citecolor=darkblue, linkcolor=darkblue, urlcolor=darkblue}

\title{MOM: Memory-Efficient Offloaded Mini-Sequence Inference for Long Context Language Models}

\author{
\begin{tabular}{cc}
\textbf{Junyang Zhang}\thanks{Equal contribution.} & \textbf{Tianyi Zhu}\footnotemark[1] \\
\normalfont California Institute of Technology & \normalfont California Institute of Technology \\
\normalfont \texttt{junyangz@caltech.edu} & \normalfont \texttt{tzhu@caltech.edu} \\
\\
\textbf{Cheng Luo} & \textbf{Anima Anandkumar}\thanks{Corresponding author. Email: anima@caltech.edu} \\
\normalfont California Institute of Technology & \normalfont California Institute of Technology \\
\normalfont \texttt{chengluo@caltech.edu} & \normalfont \texttt{anima@caltech.edu}
\end{tabular}
}

\begin{document}

\ifcolmsubmission
\linenumbers
\fi

\maketitle

\begin{abstract}

Long-context language models exhibit impressive performance but remain challenging to deploy due to high GPU memory demands during inference. We propose Memory-efficient Offloaded Mini-sequence Inference (MOM), a method that partitions critical layers into smaller "mini-sequences" and integrates seamlessly with KV cache offloading. Experiments on various Llama, Qwen, and Mistral models demonstrate that MOM reduces peak memory usage by over 50\% on average. On Meta-Llama-3.2-8B, MOM extends the maximum context length from 155k to 455k tokens on a single A100 80GB GPU, while keeping outputs identical and not compromising accuracy. MOM also maintains highly competitive throughput due to minimal computational overhead and efficient last-layer processing. Compared to traditional chunked prefill methods, MOM achieves a 35\% greater context length extension. More importantly, our method drastically reduces prefill memory consumption, eliminating it as the longstanding dominant memory bottleneck during inference. This breakthrough fundamentally changes research priorities, redirecting future efforts from prefill-stage optimizations to improving decode-stage residual KV cache efficiency.

\end{abstract}

\section{Introduction}

The Transformer architecture \citep{vaswani2017attention} revolutionized natural language processing through self-attention, enabling models to capture long-range dependencies. Despite their impact, standard Transformers have inherent limitations processing long sequences due to quadratic memory complexity—a challenge that has driven extensive research into efficient Transformer variants \citep{tay2020long} and architectures tailored for long documents like Longformer \citep{beltagy2020longformer}. Concurrently, system-level innovations such as FlashAttention \citep{dao2022flashattention, dao2023flashattention}, ZeRO \citep{rajbhandari2020zero}, Megatron-LM \citep{shoeybi2019megatron}, DeepSpeed \citep{rasley2020deepspeed}, and parameter-efficient fine-tuning methods like LoRA \citep{hu2022lora} have advanced model scalability and training efficiency.

Recently, test-time computation has gained prominence, driven by techniques like few-shot learning \citep{brown2020language}, beam search \citep{snell2024scaling}, and prompt engineering strategies such as chain-of-thought prompting \citep{wei2022chain}. These techniques shift computational demands from training to inference. Large language models like ChatGPT now dynamically expand context, highlighting the critical need for efficient GPU memory management during inference—especially when sophisticated decoding methods like beam search, lookahead search \citep{snell2024scaling}, Tree of Thoughts \citep{yao2023tree}, and Forest of Thoughts \citep{bi2024forest} significantly increase memory usage.

Consumer-grade GPUs typically have limited memory, while enterprise ones with more memory usually come at a much higher price tag. This highlights the need to optimize VRAM usage for effective performance on affordable hardware. Typically, the MLP layers dominate peak memory usage due to large intermediate activations and computational intensity. Although attention layers also contribute, optimizations such as FlashAttention, Linformer \citep{wang2020linformer}, Reformer \citep{kitaev2020reformer}, Multi-Query Attention (MQA) \citep{shazeer2019fast}, and Grouped-Query Attention (GQA) \citep{ainslie2023gqa} mitigate their impact.

Mini-Sequence Transformer (MST) \citep{Chengluo2024} leverages gradient checkpointing \citep{chen2016training} and gradient accumulation \citep{you2019large} to partition large intermediate values into smaller mini-sequences. MST significantly reduces peak GPU memory usage but is training-focused and unsuitable for efficient inference due to overhead from gradient operations. HEADINFER \citep{luo2025headinfer} further reduces GPU memory demands by employing a fine-grained, head-wise KV cache offloading strategy; however, it suffers from significant decoding speed degradation (7–8 times slower than standard LLMs).

\begin{figure}[t]
    \centering
    \includegraphics[width=0.9\linewidth, height=7cm]{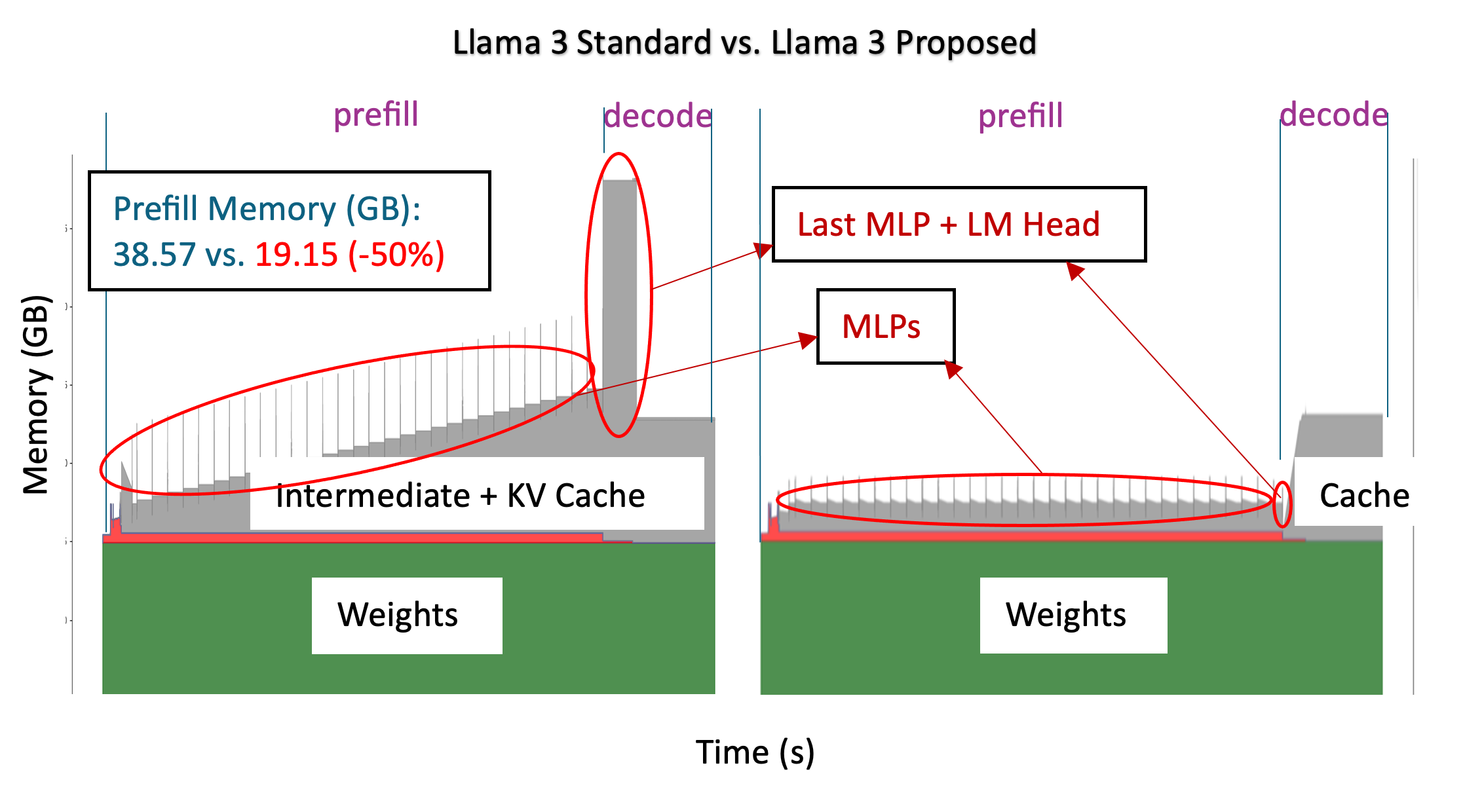}
    
    \caption{ GPU Memory Comparison of Llama 3 Standard vs. Llama 3 with MOM for a 64K Input Context.}
    \label{fig:fullcompare}  
\end{figure}

\textbf{Our Approach}: Recognizing these challenges, we propose Memory-efficient Offloaded Mini-sequence Inference (MOM), which offloads the KV cache from GPU to CPU during prefill and reloads it during decode stage, while internally partitioning the inputs to MLP layers into smaller mini-sequences and processing only one token at the final MLP and LM head to improve throughput and memory efficiency. As illustrated in Figure~\ref{fig:fullcompare}, MOM effectively eliminates prefill memory as the dominant bottleneck, shifting future research focus to the decode stage, where residual KV cache optimization becomes essential. Compared to conventional chunked prefill strategies \citep{agrawal2024taming}, which suffer from repeated forward-pass overhead, MOM processes internal mini-sequences efficiently in a single forward pass, integrating seamlessly with KV cache offloading. Also because Mini-sequence operates exclusively on the MLP and LM head and leaves the attention layers unchanged, KV cache offloading can
be seamlessly integrated with Mini-sequence.

\begin{figure}[H]
    \centering
    \includegraphics[width=0.9\linewidth]{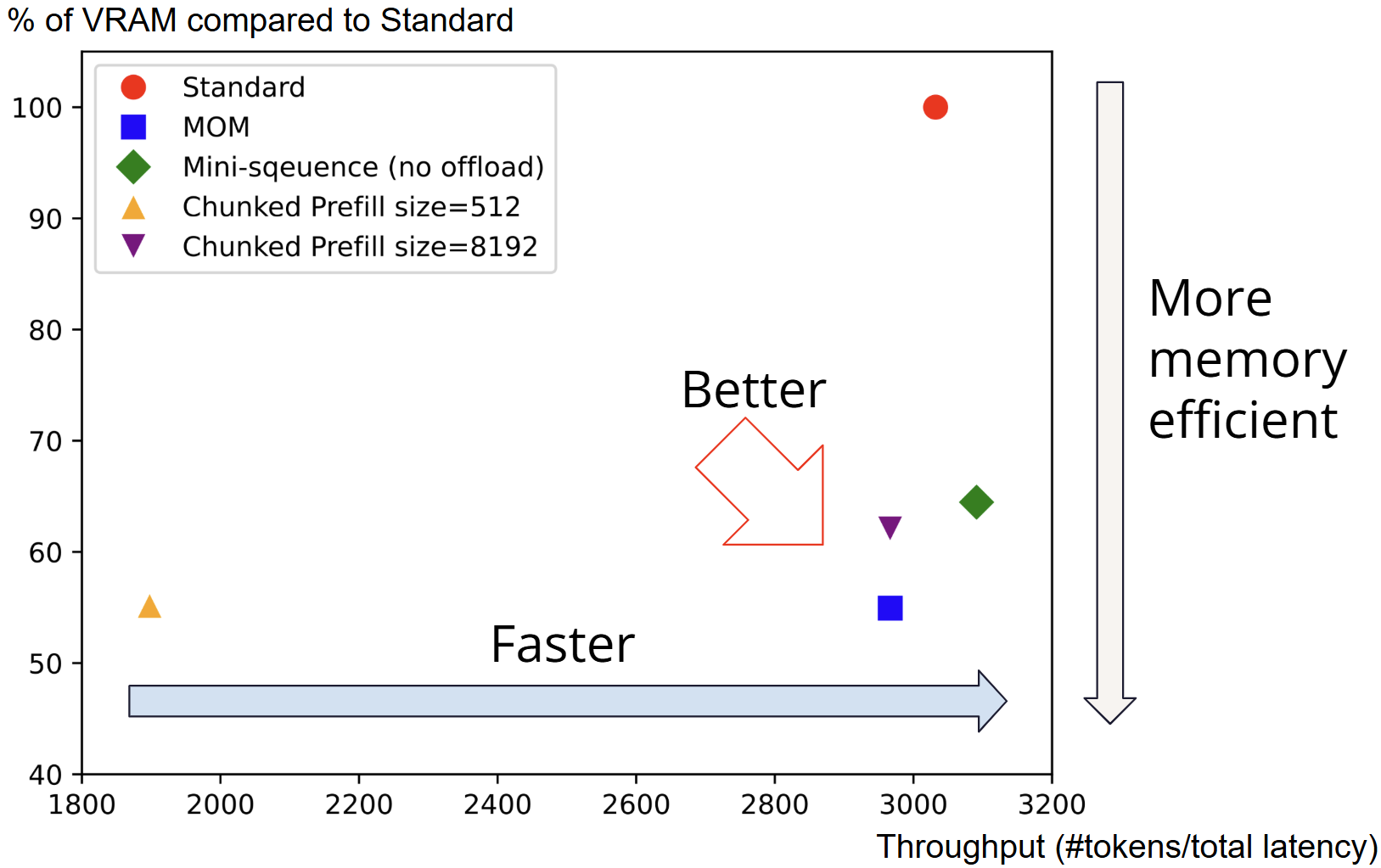}
    \caption{Memory vs. Throughput (Average of Various Input Sequence Lengths). }
    \label{fig:memory_speed_tradoff}
\end{figure}

We conduct extensive experiments on Llama \citep{touvron2023llama}, Qwen \citep{qwen2.5}, and Mistral \citep{mistral_nemo}, evaluating baseline, offloading alone, Mini-sequence alone, and combined Mini-sequence with offloading (MOM) configurations on a NVIDIA A100 80GB GPU. For example, MOM reduces Meta-Llama-3-8B peak memory usage from 72 GB to 35 GB for a 155K-token context, extending maximum context length to 455K tokens—35\% greater than chunked prefill methods. Besides, as shown in Figure~\ref{fig:memory_speed_tradoff}, its throughput degradation is minimal. Conventional chunked prefill, if combined with cache offloading, would suffer a throughput reduction of more than 75\%, making this combination extremely impractical due to data transfer overhead. Interestingly, Mini-sequence inference without offloading even improves throughput and token generation speed, due to more efficient last-layer processing, better GPU cache utilization, and reduced memory allocation overhead. We hypothesize that shorter sequence chunks fit better into GPU cache than longer sequences, enabling faster processing and thus supporting longer contexts and complex decoding without sacrificing speed.

Our contributions include:
\begin{itemize}
\item \textbf{Memory Efficiency:} MOM reduces peak GPU memory usage by over 50\%.
\item \textbf{Extended Context Length:} Extends context lengths from 155K to 455K tokens.
\item \textbf{High Throughput:} Achieves competitive token generation speeds.
\item \textbf{Mathematical Equivalence:} Preserves output content without accuracy degradation.
\item \textbf{Outperforms Chunked Prefill:} Offers 35\% longer context extension without repeated forward pass overhead.
\item \textbf{Ease of Use:} Implementation-agnostic with minimal changes required for frameworks like Hugging Face \citep{jain2022introduction}.
\end{itemize}

The implementation is available on GitHub: \url{https://github.com/TianyiZhu877/MOM}


\section{Related Work}
\textbf{KV Cache Offloading} KV cache offloading is a technique used in LLM inference to manage memory constraints when processing long sequences. Since the key-value (KV) cache stores past attention states, its size scales linearly with sequence length and can quickly exceed GPU memory capacity. Offloading moves inactive or less frequently accessed KV cache tensors to CPU memory, NVMe storage, or lower-bandwidth GPU memory, freeing up high-speed HBM (High Bandwidth Memory) for active computation. This is particularly useful for batched inference and long-context models, where keeping the entire KV cache in GPU memory would be impractical. Advanced implementations, such as PagedAttention \citep{kwon2023efficient}, further optimize this by dynamically swapping only necessary KV blocks back to GPU when needed, reducing data transfer latency. Efficient KV cache offloading allows scalable long-sequence inference without excessive memory overhead, improving overall throughput and system efficiency.

\textbf{MLP-Dominated Prefill Memory} In large language model (LLM) inference, the prefill stage dominates peak GPU memory consumption, primarily due to the MLP (feed-forward) layers rather than the attention layers \citep{kalra2023memory}. During the prefill stage, the entire input sequence is processed in parallel, requiring \(\mathcal{O}(\text{sequence\_length} \times d_{\text{model}}^2)\) memory for the MLP layers. While self-attention contributes to memory usage—particularly due to key-value (KV) cache growth in long sequences—it scales \(\mathcal{O}(\text{sequence\_length}^2 \times d_{\text{model}})\) in standard attention, which can be optimized using FlashAttention and Grouped-Query Attention (GQA). In contrast, MLP layers involve large matrix multiplications with weights that cannot be easily pruned or quantized without affecting accuracy, making them the dominant factor in peak GPU memory usage. A detailed illustration is provided in  Fig~\ref{fig:fullcompare}. As shown by Figure~\ref{fig:prefilldecode} in Appendix~\ref{sec:prefill_decode}, during the decode stage, memory usage is dominated by the KV cache rather than the MLP layers. Since only one token is processed per step, it grows linearly with sequence length and does not exceed the peak seen in prefill \citep{lienhart2024kv}. 

\textbf{Chunked Prefill} To address the MLP bottleneck in prefill stage, Chunked Prefill and its variants are widely used in academia \citep{agrawal2023sarathi} \citep{agrawal2024taming} and industry  ( by NVIDIA in TensorRT-LLM \citep{nvidia2024chunkedprefill}) to mitigate the peak memory usage in the prefill stage by splitting the input sequence into smaller chunks (see Algorithm~\ref{alg:chunked_prefill} in Appendix~\ref{sec:appendix_algo}). This allows GPUs to process smaller sections of the input, reducing intermediate memory requirements while keeping high parallelism in matrix multiplications. Chunked prefill is particularly useful for optimizing batch inference workloads, reducing VRAM spikes, and preventing out-of-memory (OOM) errors while maintaining high throughput. Similar to Mini-sequence, it reduces theoretical peak intermediate memory to \( \frac{1}{C} \) of its original size, for a chunk size of $C$.
However, unlike Mini-sequence, which only partitions the MLP layer, chunked prefill splits the entire prefill process and computes each chunk sequentially. As a result, it can result in higher latency compared to full-sequence prefill, as the overhead from multiple kernel launches and data movement may outweigh the benefits for shorter sequences.

\textbf{Mini-Sequence Transformer} The Mini-Sequence Transformer (MST) optimizes LLM training by internally partitioning input sequences into mini-sequences before each MLP layer, reducing intermediate memory usage in MLP and LM-Head layers. This method minimizes peak memory consumption while maintaining full-sequence accuracy and throughput. MST enables 12× longer sequence training without degradation, extending models like Llama3 \citep{grattafiori2024llama}, Qwen \citep{bai2023qwen}, Mistral \citep{jiang2023mistral7b} and Gemma \citep{riviere2024gemma} by 12-24×. Applying MST-like chunking to inference offers key advantages over Chunked Prefill, which splits input sequences dynamically at runtime, causing memory fragmentation, synchronization overhead, and inefficient GPU utilization. MST, by contrast, naively partitions sequences within the model architecture, reducing activation memory without extra inference-time computation. 
\vspace{-0.5cm}
\section{Method}

In this work, we propose Memory-efficient Offloaded Mini-Sequence Inference (MOM) for long context. Let $A \in \mathbb{R}^{B \times S \times d}$ denote the input sequence's representation to the MLP layer, where $B$ is the batch size (we assume $B$ = 1 in this paper), $S$ is the sequence length, and $d$ is the hidden dimension. The core idea of Mini-sequence is to partition $A$ into $M$ shorter sequences
$(A_1, A_2, \ldots, A_M)$,
where each sequence $A_i \in \mathbb{R}^{B \times N \times d}$ with $N \approx S/M$. In our inference setting, we apply Mini-sequence exclusively to the MLPs and only take the last token to feed the last MLP layer and LM-head block, leaving the attention layers unchanged so that existing optimizations such as FlashAttention and Grouped-Query Attention can continue to operate. Crucially, by decoupling from gradient computations, we can integrate offloading to move KV caches to CPU memory (or disk) when they are not actively used, thereby further reducing the GPU memory footprint.

\subsection{Mini-Sequence Processing for Inference}
\begin{figure}[ht]
    \centering
    \includegraphics[width=0.65\linewidth]{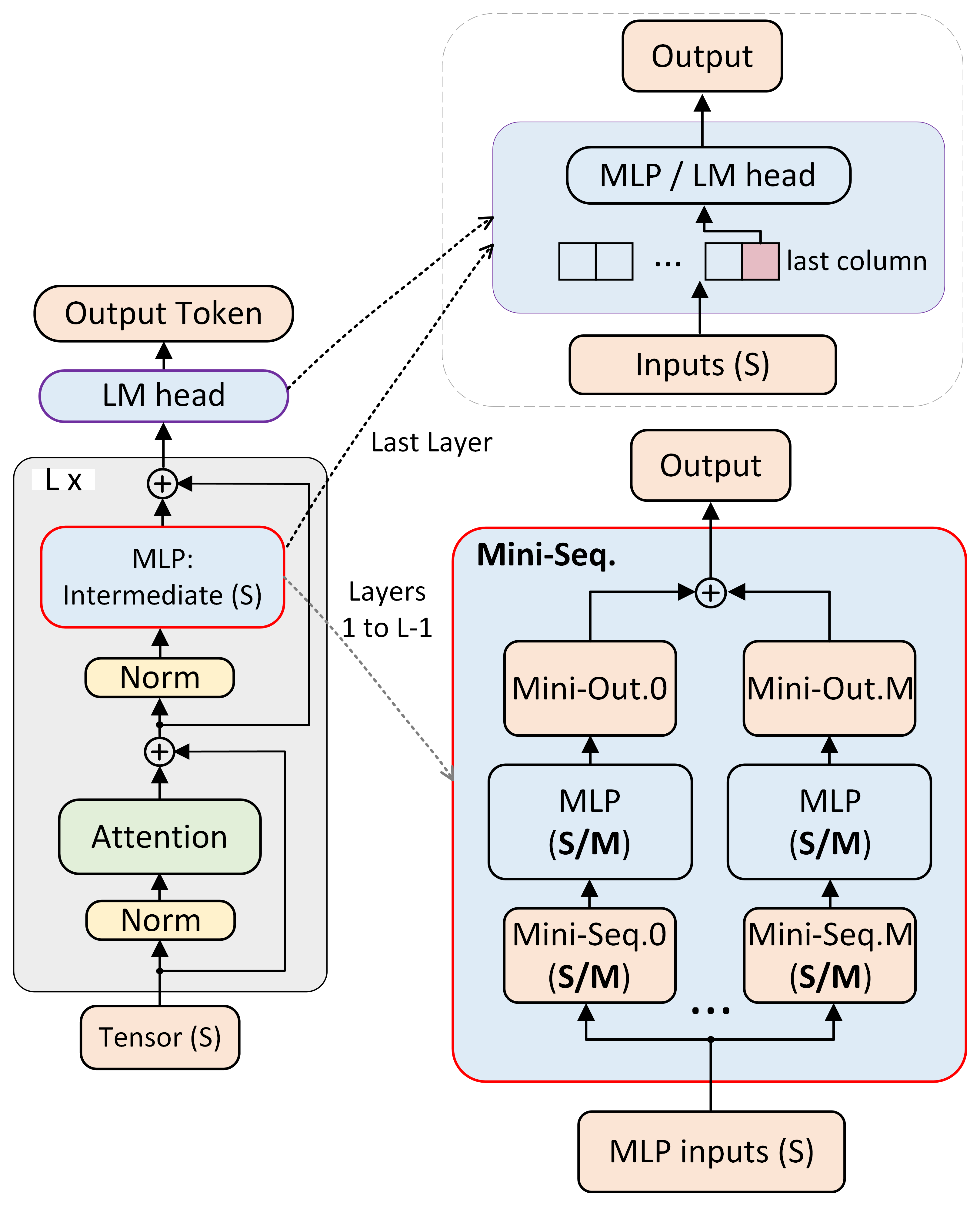}
    \caption{MOM Architecture Overview.}
    \label{fig:mst-architecture}
\end{figure}

During the prefill stage, where the entire prompt is processed to initialize the KV cache, we employ internal chunking within the MLP blocks as shown in Figure~\ref{fig:mst-architecture}. When performing autoregressive decoding, we only project the final token’s hidden state to the last MLP layer and LMHead layer to obtain the next-token logits. This is formalized in Algorithm ~\ref{alg:mst_inference}.
\vspace{-0.45cm}
\vspace{-0.6cm}
\begin{algorithm}[H] 
\caption{ Memory-efficient Offloaded Mini-Sequence Inference}
\label{alg:mst_inference}
\begin{algorithmic}
\Require Input sequence $X \in \mathbb{R}^{B \times S \times d}$, Mini-sequence size $C$, offloaded KV cache $\mathcal{K}$, feedforward layer  $MLP$, batch size $B$, sequence length $S$, and hidden dimension $d$. 
\State Compute attention layer output $A = \text{Attention}(X)$ 
    \State Update and offload KV cache to CPU: $\mathcal{K} \leftarrow \text{offload}(\mathcal{K}, A)$

\If{last MLP layer}
\State Extract last token representation: $A_{\text{last}} = A[:, -1, :]$ \Comment{Select last token's representation}
\State Compute MLP output $O_{\text{last}} = \text{MLP}(A_{\text{last}})$

    \State Compute logits: $L = \text{LLM\_Head}(O_{\text{last}})$
    \State Transfer offloaded cache back to GPU for decode stage.
    \State \Return $L$ \Comment{Return logits for the last token to start autoregressive decoding}
\Else
\State Partition $A$ into $M = \lceil S/C \rceil$ mini-sequences $\{A_i\}_{i=1}^{M}$, where each $A_i \in \mathbb{R}^{B \times N \times d}$ and $N \approx C$.
\For{$i = 1, \ldots, M$} 
    \State Compute $O_i = \text{MLP}(A_i)$ \Comment{Mini-sequence processing through MLP layers}
\EndFor
\State Concatenate outputs: $O = \text{concat}(O_1,\ldots,O_M)$.
\State \Return $O$. \Comment{Continue processing in the next transformer block}
\EndIf

\end{algorithmic}
\end{algorithm}

\subsection{KV Cache Offloading Integration}
\begin{figure}[ht]
    \centering
    \includegraphics[width=0.63\linewidth, height=3cm]{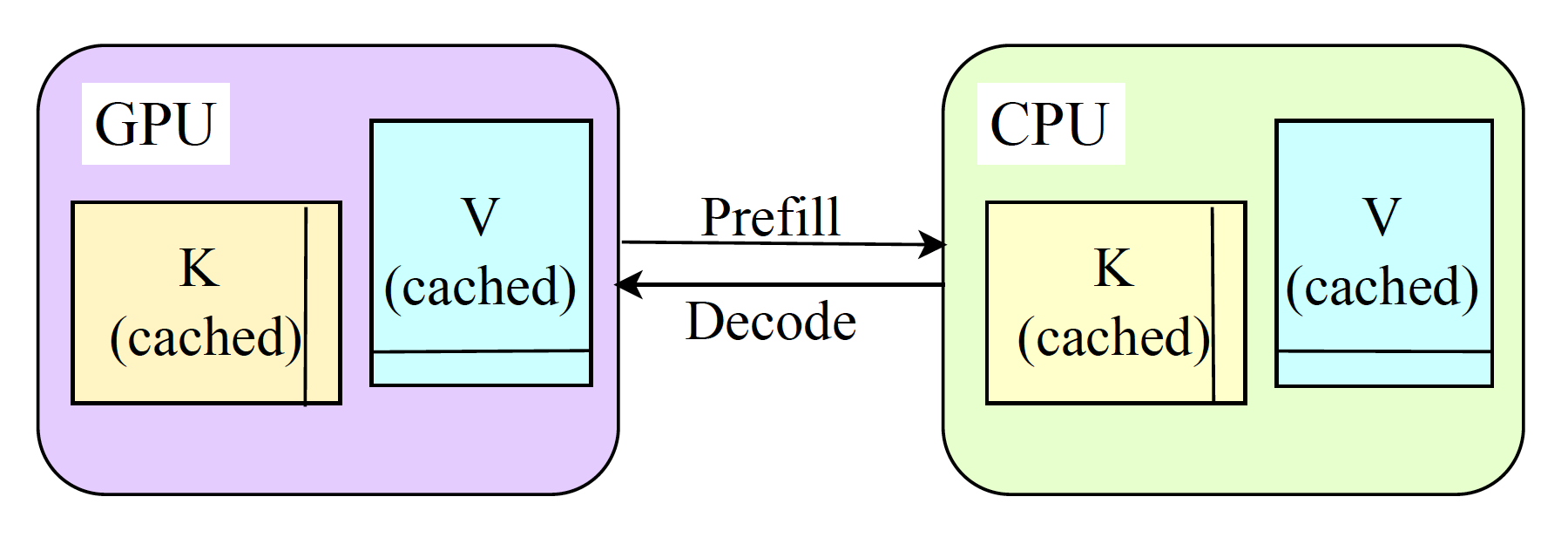}
\vspace{-0.5cm}
    \caption{ Dynamic KV Cache Transfer Between GPU and CPU in Prefill and Decode Stages.}
    \label{fig:offload-details}
\end{figure}

During inference, the Transformer relies on a KV cache to store intermediate attention states. Our method leverages existing offloading mechanisms (e.g., via Hugging Face's \texttt{transformers.cache\_utils.OffloadedCache} class) to move inactive KV cache tensors to CPU memory, as shown in Figure~\ref{fig:offload-details}. The offloading integration is dynamic: before processing mini-sequences, the corresponding KV caches are updated and offloaded automatically, ensuring that only the minimal set of tensors required for the current computation resides on GPU when the token's representations are processed by MLP layers. During the decode stage, the KV cache is reloaded back to GPU to prevent frequent cache transfer overheads in autoregressive decoding. This is detailed in Algorithm ~\ref{alg:mst_inference}.

\subsection{Analysis: Memory Efficiency of MOM}

\paragraph{Llama MLP Layer}

The Llama MLP layer utilizes a SwiGLU (Swish-Gated Linear Unit) architecture, enhancing efficiency and expressivity compared to standard Transformer feed-forward networks \citep{shazeer2020glu, touvron2023llama}. It employs three key matrices: \( W_{\text{gate}} \) (gating), \( W_{\text{up}} \) (expansion), and \( W_{\text{down}} \) (compression). Input \( X \) is first projected through \( W_{\text{gate}} \) and \( W_{\text{up}} \); the gating projection uses a Swish activation \( \text{Swish}(X W_{\text{gate}}) \), adaptively modulating feature importance \citep{ramachandran2017searching}. Its output is then multiplied element-wise with the expanded features from \( W_{\text{up}} \), which increases hidden dimension from \( d \) to \( 4d \). Finally, \( W_{\text{down}} \) compresses features back to dimension \( d \). This SwiGLU design improves information flow and parameter efficiency over traditional GELU-based Transformer MLPs \citep{hendrycks2016gaussian}.

\paragraph{Standard Transformers Without Optimization}  
Let $X \in \mathbb{R}^{S \times d}$ be the input sequence of length $S$, and hidden dimension $d$, number of transformer block layers $L$. In a standard (full-sequence) forward pass, the peak intermediate activation memory required for MLP blocks is $A_{\text{full}}$. The memory used during inference consists of model weights \( W_{\text{model}} \), the KV cache of size \( 2 \cdot S \cdot d \cdot L \), and the intermediate computation results of each layer:
\begin{itemize}
    \item For the attention mechanism, the theoretical peak intermediate memory is \( S \cdot S \), but optimized attention mechanisms such as FlashAttention \citep{dao2022flashattention, dao2023flashattention} and Memory Efficient Attention \citep{efficientattention} significantly reduce this. The peak memory usage is instead determined by the output size, which is \( S \cdot d \).  

    \item In the MLP layers, intermediate tensors \( I_{\text{up}}, I_{\text{gate}} \in \mathbb{R}^{S \times I} \) are generated, where $I \approx 4d$. Memory usage peaks at the Up-Projection hidden layer output, size \( S \cdot I\).  

    \item During inference, the LM head generates only one token at a time, requiring intermediate memory equal to the vocabulary size \( V \). 
\end{itemize}

Since intermediate memory does not persist throughout inference, the peak intermediate memory consumption is the maximum of these components. In models like Llama 3, $I$ is typically much larger than $d$, so this is dominated by the MLP layer:  
\begin{equation}
\mathcal{M}_{\text{intermediate}} = \max(S \cdot d, S \cdot I, V ) = S \cdot I
\end{equation}
Thus, the total peak memory consumption for inference is:  
\begin{equation}
\label{eq:origin_mem_model}
\mathcal{M}_{\text{total}} = W_{\text{model}} + \mathcal{M}_{\text{KV}} + \mathcal{M}_{\text{intermediate}} = W_{\text{model}} + 2 \cdot S \cdot d \cdot L + S \cdot I
\end{equation}
\paragraph{Mini-sequence Partitioning.} 
To reduce intermediate memory, $X$ gets split into $M$ mini-sequences, each of length $N \approx \frac{S}{M}$. Processing each mini-sequence independently lowers the peak intermediate memory to approximately
\begin{equation}
\label{eq:msi_mem_model}
\mathcal{M}_{\text{intermediate\_mini}} \approx \frac{\mathcal{M}_{\text{intermediate\_full}}}{M} =   \frac{S \cdot I}{M}
\end{equation}
\vspace{-0.5cm}

Assuming intermediate buffers are freed between mini-sequences. In practice, the memory required for each mini-sequence will be less than 
$\mathcal{M}_{\text{intermediate\_full}}$ but more than $\mathcal{M}_{\text{intermediate\_mini}}$ due to overlapping buffers and computational overhead.
\paragraph{Offloading}
During inference, key/value (KV) caches and other data can be offloaded to CPU/disk. Let $W_{\text{model}}$ be the model weights in GPU memory, and $O_{\text{offload}}$ the overhead for data transfers and buffers. Then, out of total GPU memory $M_{\text{max}}$, the effective memory available for MLP and LM head  during prefill stage is
\begin{equation}
\mathcal{M}_{\text{avail}} \;=\; \mathcal{M}_{\text{max}} \;-\; W_{\text{model}} \;-\; O_{\text{offload}}.
\end{equation}

\paragraph{Maximum Sequence Length}
Define $S_{\max}$ as the maximum sequence length fitting into GPU memory. As Mini-sequence reduces peak intermediate to $\mathcal{M}_{\text{intermediate\_mini}}$, we get
\begin{equation}
\label{eq:smax}
S_{\max} \;\propto\; \frac{\mathcal{M}_{\text{avail}}}{\mathcal{M}_{\text{intermediate\_mini}}}
\;=\;
\frac{\mathcal{M}_{\text{max}} \;-\; W_{\text{model}} \;-\; O_{\text{offload}}}{\mathcal{M}_{\text{intermediate\_mini}}}.
\end{equation}
As $M$ grows, $\mathcal{M}_{\text{intermediate\_mini}}$ decreases, allowing for larger $S_{\max}$. Equation~\eqref{eq:smax} shows that even with non-trivial offloading overhead, Mini-sequence reduces the intermediate memory per sequence sufficiently to handle much larger lengths without exhausting GPU resources. Hence, by lowering intermediate demands (via Mini-sequence) and storing much of the KV cache off-GPU (via offloading), we can substantially extend $S_{\max}$ under a given memory budget. Therefore, MOM can process longer sequences without exceeding GPU limits, effectively removes the prefill memory as the primary memory constraint and shifts the new peak memory bottleneck to the decode stage, dominated by the GPU-resident KV cache.

\section{Experiments}

We evaluate MOM on Llama 3.2 \citep{meta2024llama32}, a state-of-the-art large language model designed for high-quality text generation. We use the 8B size version with bfloat16 datatype on single A100 80G GPU. In the Appendix \ref{sec:other-models-large} and \ref{sec:other-hardware}, we expand our tests to include other models. The evaluation examines the combination of Mini-sequence inference and offloading, comparing it with alternative techniques such as chunked prefill. It covers input context lengths of [48000, 80000, 112000, 144000] tokens to ensure the results remain consistent.




\subsection{GPU Memory for Analysis}
We evaluated the peak VRAM usage of various models under different configurations, including with and without Mini-sequence and with and without offloading, and plotted the results across different context lengths.

\begin{figure}[H]
    \centering
\includegraphics[width=0.82\linewidth]{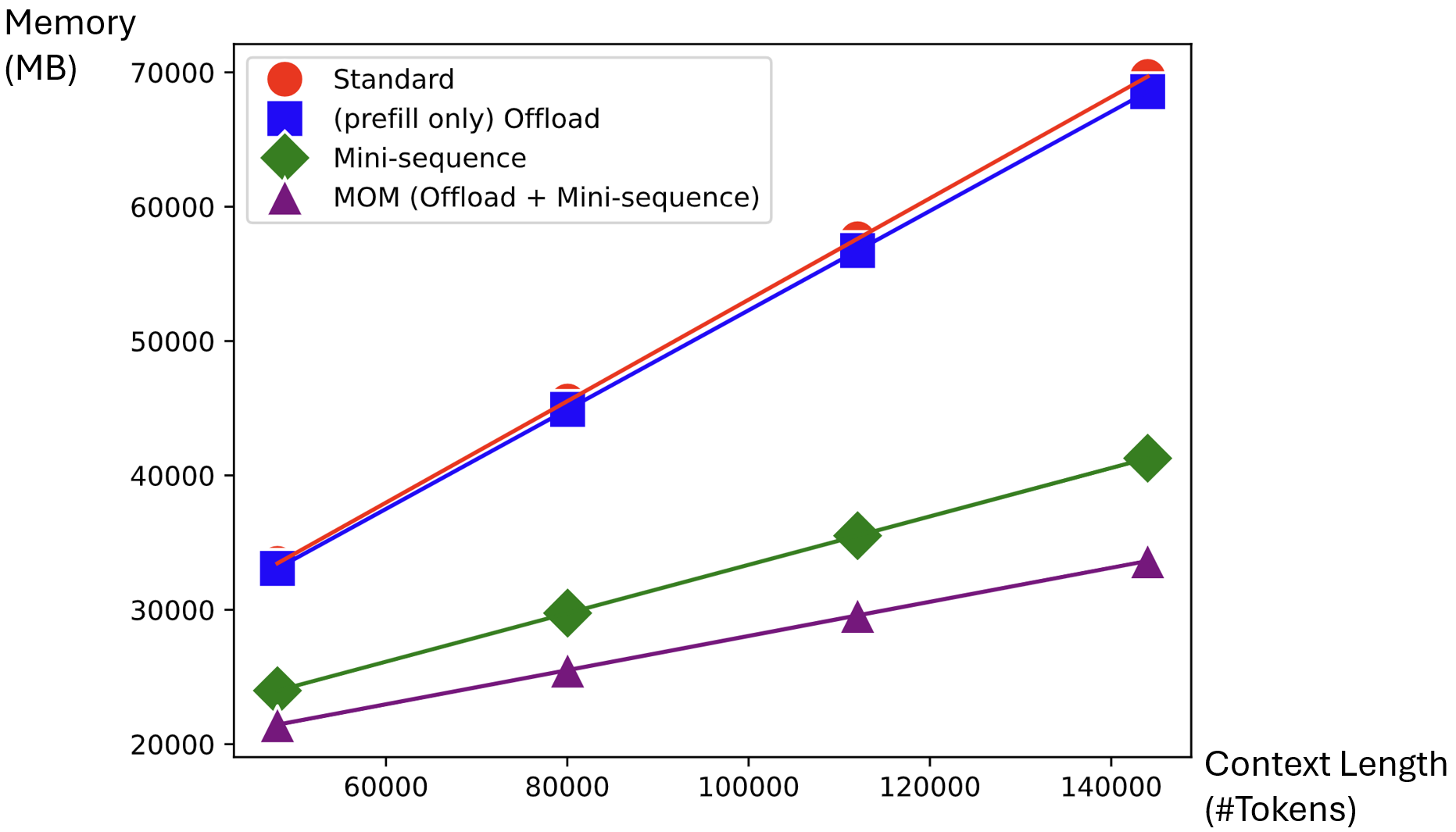}
    \caption{VRAM Comparison for Mini-sequence Inference and Offloads.}
\vspace{-0.5cm}
\label{fig:mem_curve_offload}
\end{figure}

The results in Figure \ref{fig:mem_curve_offload} align with the calculations in Equation~\eqref{eq:origin_mem_model} and Equation~\eqref{eq:msi_mem_model}, that memory use increases linearly over context length both with or without Mini-sequence. 

Mini-sequence inference has a significant impact on memory savings. And when mini-sequence inference is applied, offloading further reduces VRAM usage. However, without it, offloading alone does not lead to a substantial reduction in VRAM consumption.

As context length increases, the proportion of intermediate memory in total memory grows, since the model weight size remains unchanged. Consequently, we observe a higher percentage of memory savings with mini-sequence inference as total memory usage increases.

\vspace{-0.5cm}
\begin{table}[H]
  \caption{Memory Usage WITH Mini-sequence Divided by NO Mini-sequence on Different Offloading Schemes (\%, lower is more memory efficient)}
    \centering
  \label{table:mem_porportion}
\begin{tabular}{lrrrr}
\toprule
\textbf{Context Length} (\#Tokens): & \textbf{48000} & \textbf{80000} & \textbf{112000} & \textbf{144000} \\
\midrule
No offload & 71.682 & 65.333 & 61.643 & 59.235 \\
(prefill only) Offload & 64.834 & 56.804 & 52.146 & 49.065 \\
\bottomrule
\end{tabular}
\end{table}

\vspace{-0.5cm}

Table \ref{table:mem_porportion} also indicates that the percentage of memory savings from enabling Mini-sequence is higher when combined with offloading. This is because offloading primarily reduces KV cache or weight size, making intermediate memory—which Mini-sequence optimizes—a larger proportion of the total memory.

\subsection{Maximium Input Context Length Extension for Different Methods} 

\begin{figure}[H]
    \centering
    \includegraphics[width=0.9\linewidth]{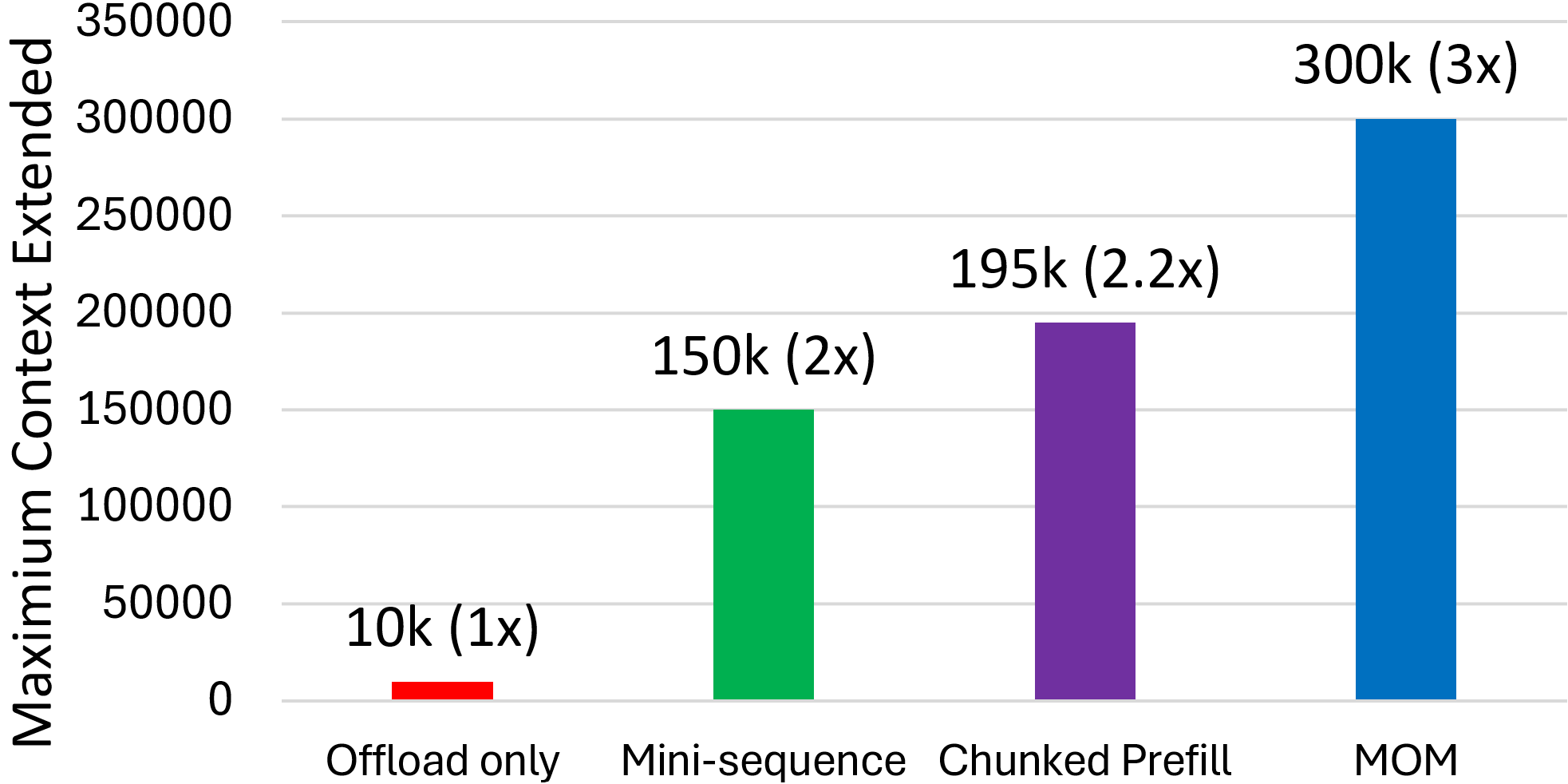}
\vspace{-0.2cm}
    \caption{Maximium Number of Context Tokens Extended from Standard Llama3.2.}
    \label{fig:max_context_experiment}
\end{figure}

We tested the maximum context length that fits into an A100-80GB GPU using different methods before encountering an Out of Memory (OOM) error. Overall, MOM outperforms all other methods, expanding the maximum context length from 155,000 tokens in the unoptimized standard model to 455,000 tokens—nearly a threefold improvement as shown in Figure~\ref{fig:max_context_experiment}.

\vspace{-0.5cm}
\subsection{Inference Speed Comparison}
\vspace{-0.6cm}
\begin{table}[H]
  \caption{Total Inference Latency (s, lower is faster)}
    \centering
  \label{latency_table}
\begin{tabular}{lrrrr}
\toprule
\textbf{Context Length} (\#Tokens): & \textbf{48000} & \textbf{80000} & \textbf{112000} & \textbf{144000} \\
\midrule
Standard & 13.971 & 23.837 & 36.592 & 52.274 \\
(prefill only) Offload & 14.693 & 25.542 & 38.538 & 56.559 \\
\textbf{Mini-sequence} & \textbf{13.536} & \textbf{23.556} & \textbf{36.160} & \textbf{51.249} \\
MOM (Mini-sequence + offload) & 14.520 & 25.020 & 38.180 & 53.417 \\
Chunked Prefill size=8192 & 14.057 & 24.515 & 37.666 & 53.247 \\
\bottomrule
\end{tabular}
\end{table}
\vspace{-0.2cm}
All the methods discussed in this section, including mini-sequence inference, offloading and chunked prefill (with 8192 chunck size), have minimal impact on speed. In Table \ref{latency_table}, we tested them across different context lengths, measuring speed by forcing the model to generate a fixed output of 200 tokens at a time and recording the total runtime for both the prefill and decoding stages. A more  detailed breakdown of prefill and decoding rate could be found in appendix \ref{sec:appen-speed}.

\subsection{Memory Speed Trade-off}
To evaluate how each optimization technique balances memory usage and speed, we measure their average memory consumption (as a percentage of the unoptimized Standard model) and throughput across multiple trials with context lengths. These results are then plotted on a scatter graph. Methods positioned closer to the bottom-right corner are generally more optimized, indicating greater memory savings with higher inference throughput. Notably, Figure \ref{fig:memory_speed_tradoff} shows that MOM appears closer to the lower-right corner, suggesting that it achieves better memory efficiency with minimal trade-offs in speed.

\subsection{Accuracy}
\textbf{Logit Equivalence Test} To validate that MOM has no effect on accuracy, we first tested random input sequences on both MOM and the standard model, comparing the output logits, which were identical.

\begin{figure}[H]
    \centering
    \begin{minipage}{0.44\textwidth}
        \centering
        \includegraphics[width=\linewidth]{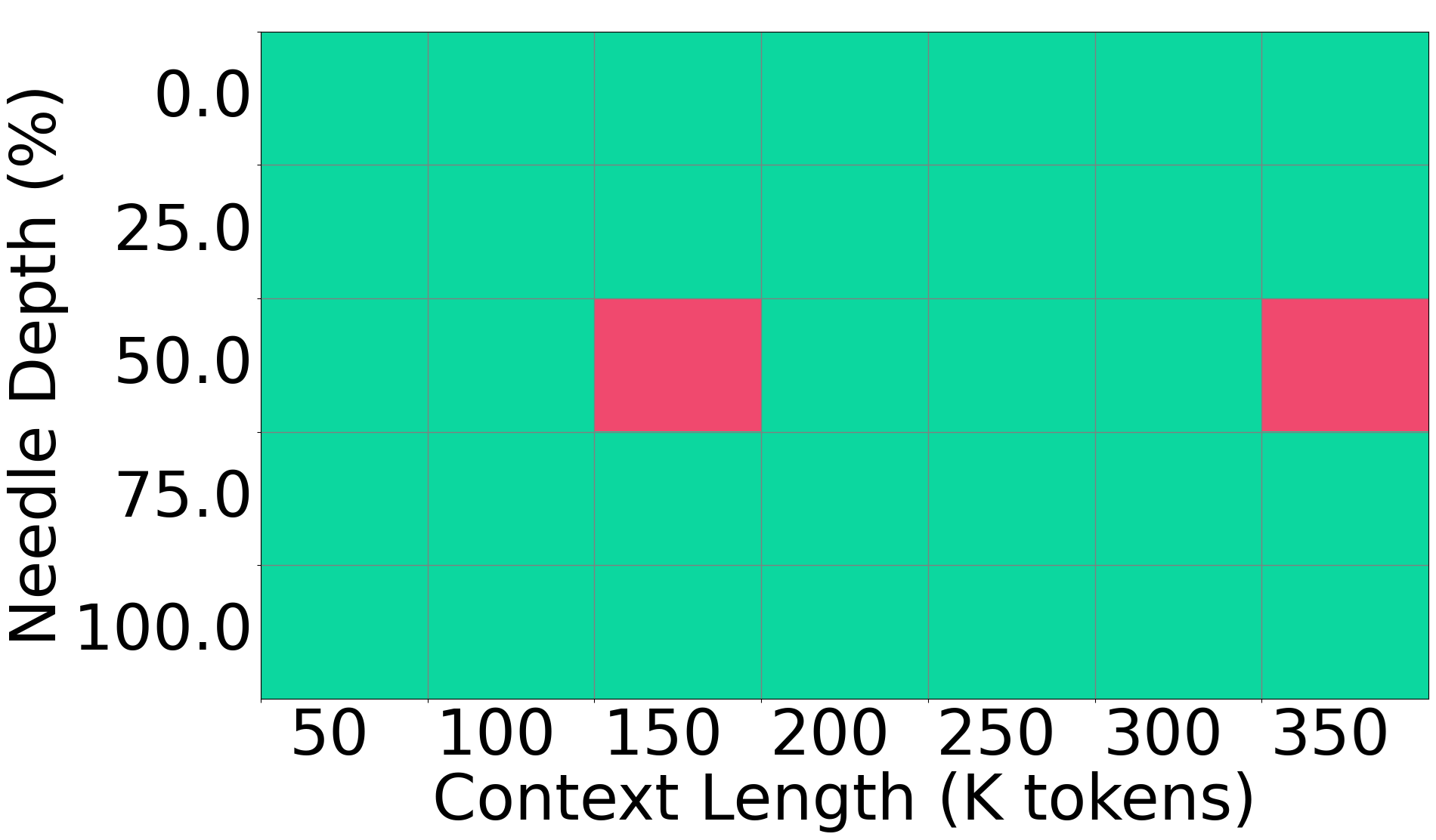}
\vspace{-0.5cm}
        \caption{\centering  Needle Test Accuracy Scores for MOM}
        \label{fig:accuracy_MOM}
    \end{minipage}
    \begin{minipage}{0.54\textwidth}
\vspace{-0.034cm}
        \centering
        \includegraphics[width=\linewidth]{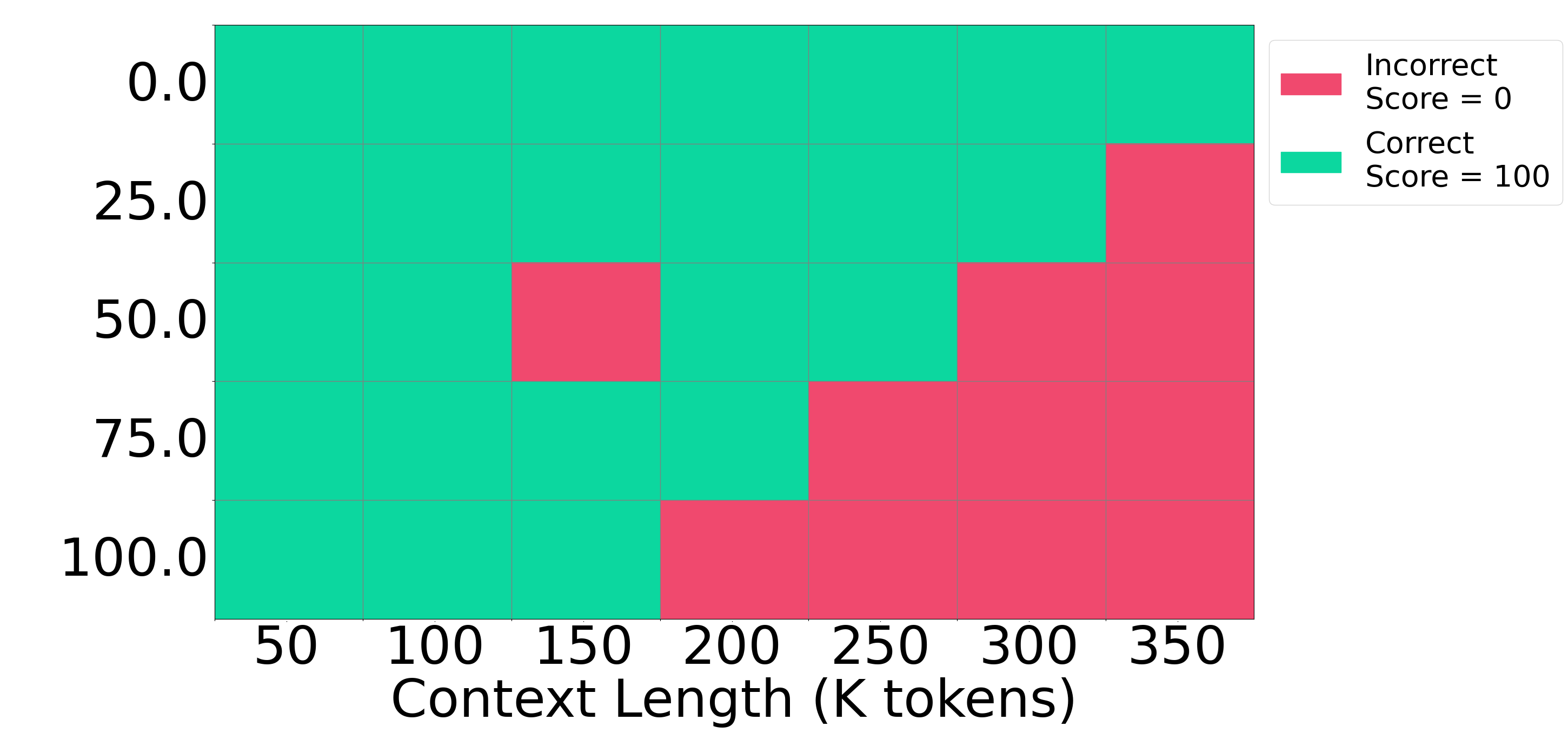}
\vspace{-0.5cm}
        \caption{\centering  Needle Test Accuracy Scores for Standard Llama3.2-8B}
        \label{fig:accuracy_standard}
    \end{minipage}
\end{figure}

\textbf{Needle Test} We evaluated the model's ability to retrieve a specific detail ("Mary's favorite number is 43251") embedded within a long, unrelated text at varying depths ($needle\ depth$). Accuracy was binary (100 if correct, 0 otherwise). As shown in Figure \ref{fig:accuracy_standard}, the standard model failed when $needle\ depth \times context\ length > 150000$ due to GPU memory constraints causing text truncation. In contrast, MOM (Figure \ref{fig:accuracy_MOM}) handled extended contexts without truncation. Occasional incorrect responses appeared similarly in both models, indicating no accuracy degradation from MOM.

 \vspace{-0.2cm}
\vspace{-0.3cm}

\section{Future Works}
\paragraph{Optimizing Integration with Other Inference Frameworks}
Beyond Hugging Face, large language model inference for individuals and small businesses is often performed using frameworks like vLLM \citep{vllm} or sglang \citep{sglang}. While the MOM mechanism is compatible with these frameworks, not all inference processes may be fully optimized or seamlessly integrated. A deeper investigation into their inference mechanisms is needed to ensure optimal performance and compatibility across different implementations.

\paragraph{Optimizing KV Cache During Inference}
Our method has significantly optimized memory usage during the prefill stage , bringing it close to optimal (Figure \ref{fig:fullcompare}). Memory consumption is now dominated by the KV cache during decoding stage, presenting an opportunity for further improvement. Future research on KV cache compression techniques for the decoding stage could complement our method, allowing for even greater memory efficiency.

\section*{Acknowledgment}
We thank Caltech CS165 support.
A. Anandkumar is supported by the Bren named chair professorship, Schmidt AI 2050 senior fellowship, ONR (MURI grant N00014-18-12624).

\section*{Ethics Statement}

Our Memory-efficient Offloaded Mini-Sequence Transformer (MOM) addresses GPU memory efficiency and computational performance for inference tasks. While MOM itself does not inherently introduce ethical concerns, the increased accessibility and efficiency of large language models enabled by our approach could amplify societal impacts, including existing biases present in the underlying datasets. We encourage practitioners adopting MOM to follow responsible AI practices, such as bias monitoring, fairness evaluations, transparency, and privacy preservation, particularly when deploying models in sensitive contexts. All experimental procedures in this work adhere strictly to ethical standards, without involving human subjects or private data.

\bibliographystyle{colm2025_conference}

\appendix
\section{GPU Memory Usage During Inference}
\label{sec:prefill_decode}

\begin{figure}[H]
    \centering
    \includegraphics[width=0.8\linewidth, height=8cm]{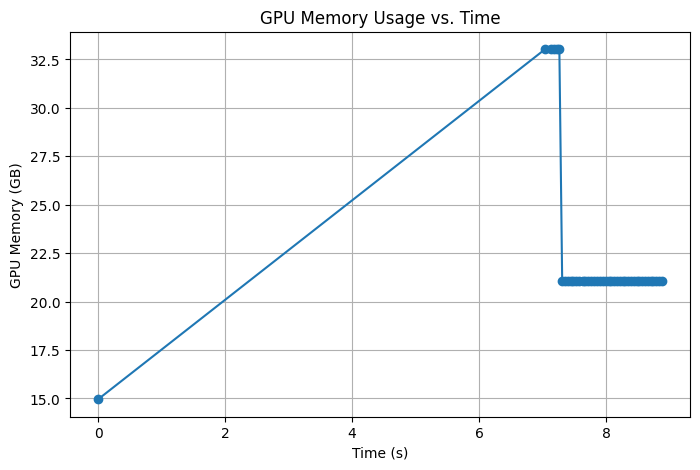}
    \caption{GPU Memory Usage During Inference: starting from the second datapoint, each datapoint represents the memory usage when generating a new token. The memory peaks before generating the first token and drops significantly during decode stage.}
    \label{fig:prefilldecode}
\end{figure}

During LLM inference, the prefill stage—where the entire input sequence is processed at once—dominates GPU memory usage due to the storage of intermediate activations and key-value (KV) cache across all tokens. In contrast, the decode stage generates output token by token, reusing the KV cache from previous steps, which results in significantly lower memory consumption as the model only processes one token at a time, as indicated in Figure \ref{fig:prefilldecode}.

\section{Basic Chunked Prefill Algorithm}
\label{sec:appendix_algo}

Chunked prefill is an alternative technique for reducing inference memory by splitting the context into smaller chunks during the prefill stage. While more complex implementations can also improve computational speed, we compare it with the simplest version (See Algorithm \ref{alg:chunked_prefill}), which is primarily designed to reduce memory usage. 

\begin{algorithm}[H]
\caption{Basic Chunked Prefill}
\label{alg:chunked_prefill}
\begin{algorithmic}
\Require Input sequence $X \in \mathbb{R}^{B \times S \times d}$, chunk size $C$, large language model $M$

\State Initialize empty key-value cache $K$
\State Split $X$ into chunks: $X^{(1)}, X^{(2)}, \dots, X^{( \lceil S/C \rceil )}$ where each $X^{(i)} \in \mathbb{R}^{B \times C \times d}$ has at most $C$ tokens
\For{each chunk $X^{(i)}$}
    \State Compute model $Output^{(i)} = M(X^{(i)}, K)$
    \State Extract and store key-value pairs in cache: $K \gets K \cup \text{KV}(Output^{(i)})$
\EndFor
\State Proceed with normal autoregressive decoding using cached $K$
\end{algorithmic}
\end{algorithm}

\section{Breakdown of Prefill and Decoding Speed of Different Methods}
\label{sec:appen-speed}

Inference in a transformer-based language model consists of prefilling and decoding stages.

\paragraph{Prefilling} This phase processes the input context before generating the first token, during which users experience a delay. This is know as the TTFT (Time to Fisrt Token), and measured for the methods discussed.
\begin{table}[H]
\caption{Time to Fisrt Token (s, lower is faster)}
  \centering
\begin{tabular}{lrrrr}
\toprule
Context Length (\#Tokens): & 48000 & 80000 & 112000 & 144000 \\
\midrule
Standard & 6.194 & 12.982 & 22.527 & 34.907 \\
(prefill only) Offload & 6.869 & 14.649 & 24.458 & 39.262 \\
Mini-sequence & 5.767 & 12.668 & 22.091 & 33.989 \\
MOM & 6.756 & 15.109 & 24.037 & 37.284 \\
Chunked Prefill size=512 & 10.526 & 24.318 & 45.321 & 72.706 \\
Chunked Prefill size=8192 & 6.286 & 13.579 & 23.530 & 35.851 \\
\bottomrule
\end{tabular}
\end{table}

The chunked prefill method splits the context into smaller chunks to reduce memory usage, but excessively small chunks significantly increase prefilling time. To balance efficiency and speed, a chunk size of 8,192 tokens is chosen in this study.

\paragraph{Decoding} After the first token is generated, the model produces subsequent tokens autoregressively at the measurable rate. No significant speed drop is observed across different methods in this stage.

\begin{table}[H]
\caption{Decode Speed,  Mini-sequence vs. Chunked Prefill (Tokens/s, higher is faster)}
  \centering
\begin{tabular}{lrrrr}
\toprule
Context Length (\#Tokens): & 48000 & 80000 & 112000 & 144000 \\
\midrule
Standard & 25.804 & 18.448 & 14.263 & 11.630 \\
(prefill only) Offload & 25.854 & 18.369 & 14.272 & 11.588 \\
Mini-sequence & 25.806 & 18.457 & 14.279 & 11.607 \\
MOM & 25.712 & 18.455 & 14.275 & 11.600 \\
Chunked Prefill size=512 & 25.837 & 18.452 & 14.276 & 11.606 \\
Chunked Prefill size=8192 & 25.868 & 18.379 & 14.220 & 11.555 \\
\bottomrule
\end{tabular}
\end{table}

\section{Testing Other LLM Models besides Llama}
\label{sec:other-models-large}

To ensure the results generalize well, we tested MOM on additional models, including Qwen2.5-7B \citep{qwen2.5} and Mistral NeMo (12B) \citep{mistral_nemo}, analyzing their speed vs. memory trade-off and comparing them with other optimization methods.

\begin{figure}[H]
    \centering
    \begin{minipage}{0.49\textwidth}
        \centering
        \includegraphics[width=\linewidth]{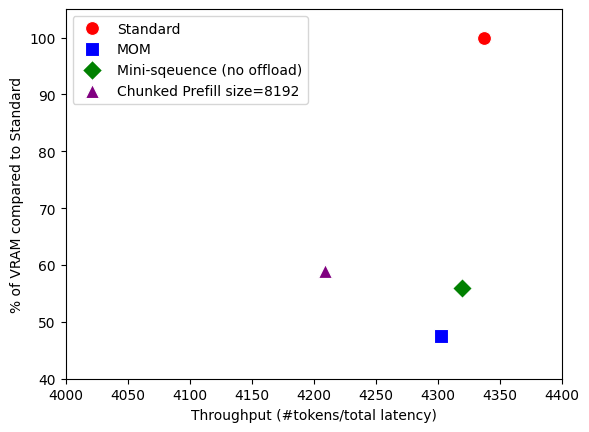}
        \caption{\centering  Memory Use vs. Throughput, Qwen2.5-7B}

    \end{minipage}
    \hfill
    \begin{minipage}{0.49\textwidth}
        \centering
        \includegraphics[width=\linewidth]{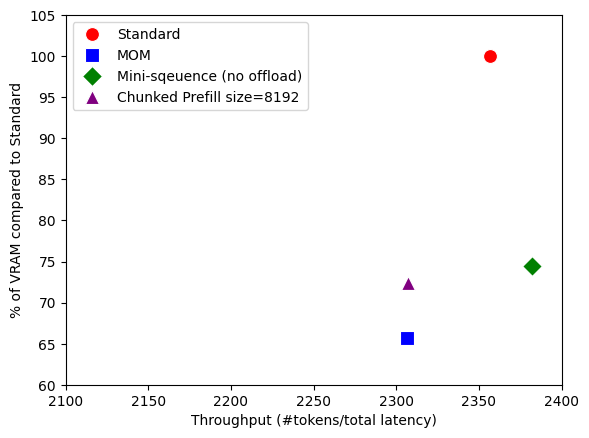}
        \caption{\centering  Memory Use vs. Throughput,      Mistral NeMo}
    \end{minipage}
\end{figure}

The results align with our findings on Llama 3.2, confirming that MOM achieves the best memory usage optimization with minimal speed overhead.

\section{Testing on Different Hardware Setup and with Quantization}
\label{sec:other-hardware}
In practice, most individual users perform inference on consumer-grade hardware with quantization. To reflect this, we include tests on an RTX 4080 mobile 12GB GPU, using bitsandbytes \citep{bitsandbytes} 4-bit quantization. Due to VRAM limitations, we tested with context lengths of [16,000, 20,000, 24,000] tokens. 

\begin{figure}[H]
    \centering
    \begin{minipage}{0.49\textwidth}
        \centering
        \includegraphics[width=\linewidth]{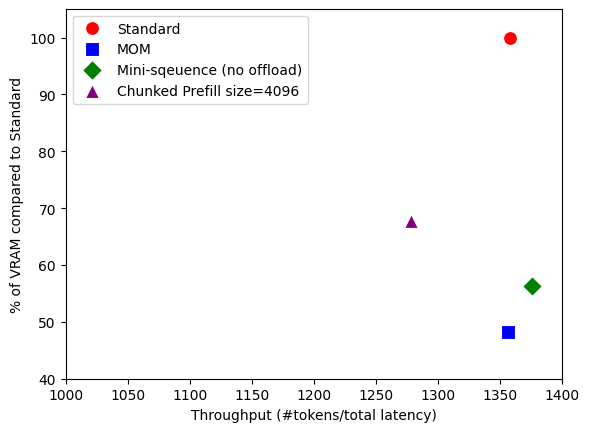}
        \caption{\centering  Memory Use vs. Throughput, Llama3.2-3B}

    \end{minipage}
    \hfill
    \begin{minipage}{0.49\textwidth}
        \centering
        \includegraphics[width=\linewidth]{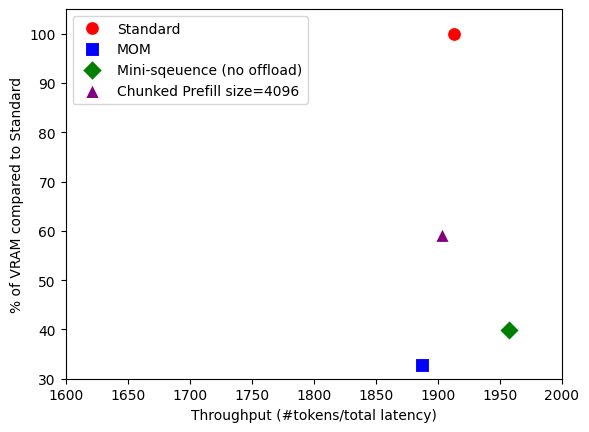}
        \caption{\centering  Memory Use vs. Throughput,  Qwen2.5-3B}
    \end{minipage}
\end{figure}

The results align with our findings with A100 GPU, reinforcing the effectiveness of MOM across different environments and practical setups.

\end{document}